\documentclass{article}

\usepackage{microtype}
\usepackage{graphicx}
\usepackage{subfigure}
\usepackage{booktabs} 

\usepackage{hyperref}

\usepackage[accepted]{icml2024}

\usepackage{amsmath}
\usepackage{amssymb}
\usepackage{mathtools}
\usepackage{amsthm}

\usepackage[capitalize,noabbrev]{cleveref}

\theoremstyle{plain}

\theoremstyle{definition}

\theoremstyle{remark}

\usepackage[textsize=tiny]{todonotes}

\usepackage{multirow}

\icmltitlerunning{Submission and Formatting Instructions for ICML 2024}

\begin{document}
	
	\twocolumn[
	\icmltitle{A Channel-ensemble Approach: Unbiased and Low-variance Pseudo-labels \\ is Critical for Semi-supervised Classification}
	
	
	
	
	\begin{icmlauthorlist}
		\icmlauthor{Jiaqi Wu}{bjut}
		\icmlauthor{Junbiao Pang}{bjut}
		\icmlauthor{Baochang Zhang}{bh}
		\icmlauthor{Qingming Huang}{ucas}
	\end{icmlauthorlist}
	
	\icmlaffiliation{bjut}{Beijing University of Technology, Beijing, China}
	\icmlaffiliation{bh}{Beihang University, Beijing, China}
	\icmlaffiliation{ucas}{University of Chinese Academy of Sciences, Beijing, China}
	
	\icmlcorrespondingauthor{Junbiao Pang}{Junbiao\_pang@bjut.edu.cn}
	
	\icmlkeywords{Machine Learning, ICML}
	
	\vskip 0.3in
	]
	
	
	
	\printAffiliationsAndNotice{\icmlEqualContribution} 
	
	\begin{abstract}
		Semi-supervised learning (SSL) is a practical challenge in computer vision. Pseudo-label (PL) methods, e.g., FixMatch and FreeMatch, obtain the State Of The Art (SOTA) performances in SSL. These approaches employ a threshold-to-pseudo-label (T2L) process to generate PLs by truncating the confidence scores of unlabeled data predicted by the self-training method. However, self-trained models typically yield biased and high-variance predictions, especially in the scenarios when a little labeled data are supplied. To address this issue, we propose a lightweight channel-based ensemble method to effectively consolidate multiple inferior PLs into the theoretically guaranteed unbiased and low-variance one. Importantly, our approach can be readily extended to \emph{any} SSL framework, such as FixMatch or FreeMatch. Experimental results demonstrate that our method significantly outperforms state-of-the-art techniques on CIFAR10/100 in terms of effectiveness and efficiency.
	\end{abstract}
	
	\section{Introduction}
	\label{sec:Intro}
	
	Classification is a crucial task in computer vision and has achieved the remarkable success in various real-world applications, especially in supervised learning scenarios. However, obtaining sufficient labeled data poses challenges, and the manual annotation of samples is the time-consuming process. For example, the medical-related tasks often require the extensive analysis of the same sample by multiple human experts. As a result, Semi-Supervised Learning (SSL) \cite{arazo2020pseudo, lee2013pseudo, tarvainen2017mean, zheng2021rectifying, berthelot2019remixmatch, berthelot2019mixmatch, laine2016temporal}, which utilizes abundant unlabeled data to improve performance with limited labeled samples, has gained the significant attention.
	
	\begin{figure}[t!]
		\begin{minipage}{0.49\linewidth}
			\vspace{3pt}
			\includegraphics[width=\textwidth]{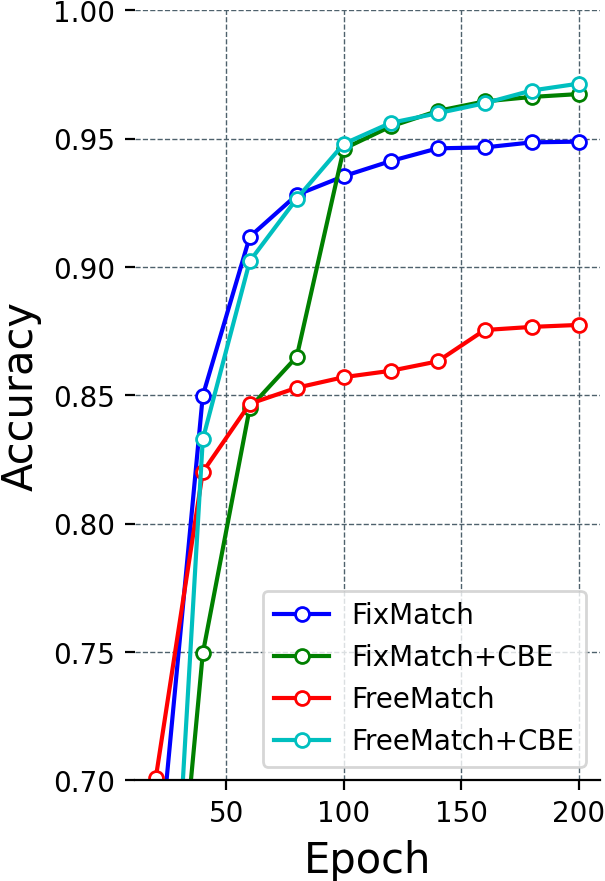}
			\centerline{(a) Accuracy}
		\end{minipage}
		\begin{minipage}{0.48\linewidth}
			\vspace{3pt}
			\includegraphics[width=\textwidth]{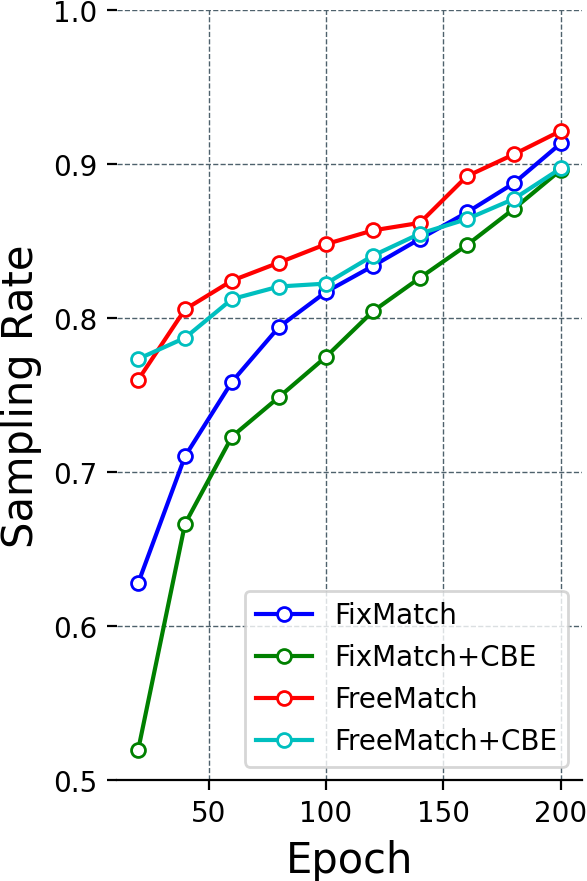}
			\centerline{(b) Sampling rate}
		\end{minipage}
		\caption{Comparison the sampling rate (SR) and accuracy of the generated PLs between FixMatch/FreeMatch and ours on CIFAR-10. The proposed method significantly improves the quality of pseudo-labels, maintaining a low SR.}
		\label{fig:PLs_Quality}
	\end{figure}

	The Pseudo Labeling (PL) method \cite{sohn2020fixmatch, zhang2021flexmatch, wang2022freematch} has shown excellent performances in SSL, employing a threshold-based filtering mechanism to covert the confidence scores predicted by a model into the pseudo-labels of samples. This process is also known as the Self-Training (ST) process. The T2L process is feasible if the confidence score of a sample can be treated as the reliable prediction. 
	
	However, as shown in Fig. \ref{fig:PLs_Quality}, the generation of the highly reliable PLs remains the challenges for both FixMatch and FreeMatch. Because a high confidence score does not means that it is a reliable prediction, especially, for a wrong prediction. Besides, due to the nature of the ST process, the PL method would hardly consistently improve performances when: 1) the discriminative ability of the a model is too weak to obtain the low bias and low variance PLs, especially when a few labeled samples are used; and 2) the wrong PLs would be accumulated over time, resulting in the high biased and high-variance predictions. 
	
	Motivated by the above two observations, we argue that the high-quality PLs should be unbiased and low-variance, which can reduce the accumulated prediction error of a model during the ST process. The empirical evidence from the classical machine learning methods( e.g. random forest~\cite{dasarathy1979composite, hansen1990neural} ) suggest that ensemble the low-biased and high-variance classifiers leads to a better prediction performance than a single classifier. However, among the current deep ensemble methods, the Temporal Ensemble (TE)~\cite{laine2016temporal} lacks the theoretical guidance and is difficult to generalize to any scenarios. In addition, the Model Ensemble (ME)~\cite{ke2019dual, tang2021humble, wu2023generating} is impractical for the practical SSL scenarios since the multiple networks are stored in memory and the corresponding training time is unbearably long.
	
	In this paper, we propose a lightweight, efficient, plug-and-play ensemble approach, named Channel-Based Ensemble (CBE), to obtain the unbiased and low-variance PLs. The CBE provides the ensemble structure with a nearly negligible computational-cost. In addition, we propose a Low Bias (LB) loss function to maximize the feature un-correlations among the multiple heads of CBE, thereby reducing the prediction bias. We further propose a Low Variance (LV) loss function to reduce the variance of the predicted distribution of the unlabeled data by utilizing the ground truth of the labeled data as constraints. Our approach can be easily extended to \emph{any} SSL framework, e.g., FixMatch, FreeMatch, and achieves better performances than the State-Of-The-Art (SOTA) techniques on CIFAR10/100. Our contributions are as follows:
	\begin{itemize}
		\item The SOTA methods either focus on the data augmentation or the selection of PLs, ignoring the characteristics of PLs during the ST process. In this paper, to the best of our knowledge, we firstly discuss the PL generation problem and propose a lightweight, efficient ensemble method based on Chebyshev constraints for generating the unbiased and low-variance PLs in SSL. 
		\item Our approach outperforms the SOTA methods on CIFAR10/100 with a fewer epochs, and can be easily extended to the other SSL frameworks like FixMatch, FreeMatch. 
	\end{itemize}
	
	\section{Related Works}
	\label{sec:Related}
	
	The mainstream approaches in SSL (i.e., the PL methods and the Consistency Regularization (CR) methods) primarily focus on the supervision signals or the threshold setting strategies, but overlook the quality of the PLs.
	
	PL methods~\cite{arazo2020pseudo, lee2013pseudo, sohn2020fixmatch, zhang2021flexmatch, wang2022freematch} reuse the PLs generated by a model to recursively train itself. The ST models accumulate their own errors during the training process, leading to the biased and high-variance predictions, especially when the amount of labeled data is relatively small, e.g. 40 labeled samples for CIFAR-10. To handle this problem, \cite{sohn2020fixmatch} constructs the weak-strong mechanism that uses the predictions of the weak augmentation samples to supervise the predictions of the strong augmentation samples. Moreover, by constructing the threshold-setting mechanisms, \cite{sohn2020fixmatch, zhang2021flexmatch} select the high-confidence predictions from the generated one-hot predictions as PLs for the unsupervised learning loss. However, these PL methods barely solve the problem of the biased and the high-variance predictions since the PLs are generated from the biased models in nature.
	
	CR methods~\cite{sajjadi2016regularization, laine2016temporal, tarvainen2017mean, miyato2018virtual, bachman2014learning, ke2019dual} encourage models to produce approximate predictions for the different perturbations from the same sample. CR methods mainly push the decision surface of the model away from the sample center through the consistency loss of unlabeled samples. It does not inherently address the low-bias and high-variance of the model to generate PLs for samples.
	
	\subsection{Strategies to reduce bias and variance}
	
	\textbf{Reducing bias.} The current approaches for mitigating the prediction bias in classification  encompass the data resampling-based \cite{qraitem2023bias} and the feature representation-based\cite{kang2020exploring} bias correction methods. The underlying principle of the former is to adjust the weights or replicate samples within the training data, thereby ensuring a more balanced sampling distribution. The latter focuses on enhancing the feature extractor to extract discriminative features that are independent of specific categories. However, these methods are not suitable for SSL scenarios.
	
	\textbf{Reducing variance.} Currently, addressing prediction variance in classification problems primarily involves the regularization-based methods \cite{moradi2020survey}.  Regularization effectively mitigates the prediction variance by constraining  the model complexity to prevent the over-fitting problem. Additionally, regularization approaches (e.g., Dropout and weight sharing) reduce the model's sensitivity to the training data by randomly deactivating certain neurons or enforcing weight sharing within the network. However, how to utilize these methods for SSL is still an open problem.
	
	\subsection{Ensemble Learning}\label{sec:related-ensemble}
	
	One way to solve the high bias and the high variance problem is through ensemble learning, which benefits from the diversity of ensemble frameworks. Existing methods for ensemble learning include Model Ensemble (ME), Temporal Ensemble (TE), and Multi-head Ensemble (MHE). 
	
	\begin{figure}[t!]
		\centering 
		\begin{minipage}{0.8\linewidth}
			\vspace{3pt}
			\includegraphics[width=1.0\textwidth]{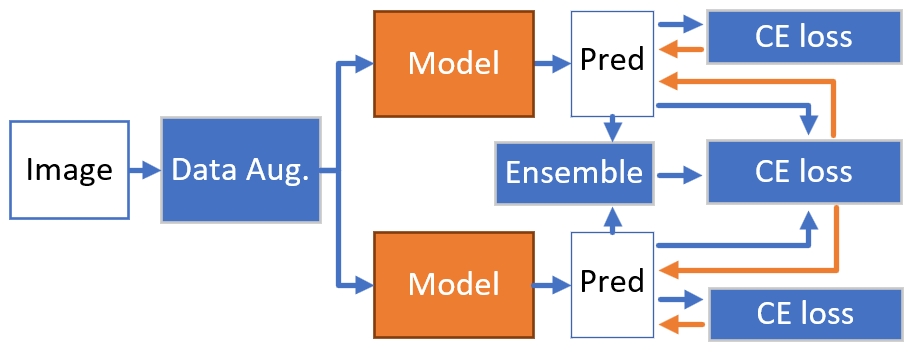}
			\centerline{(a) Model Ensemble (ME)}
		\end{minipage}
		\begin{minipage}{0.8\linewidth}
			\vspace{3pt}
			\includegraphics[width=1.0\textwidth]{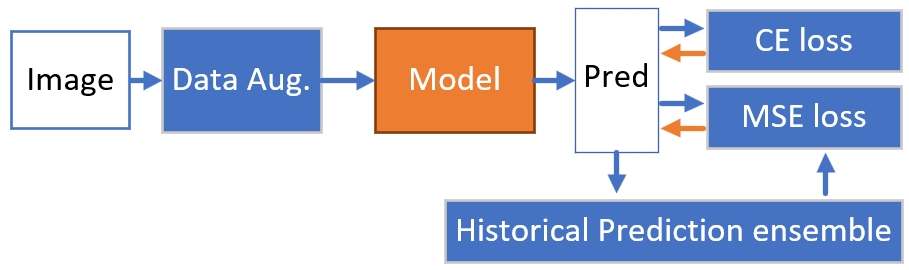}
			\centerline{(b) Temporal Ensemble (TE)}
		\end{minipage}
		\begin{minipage}{0.8\linewidth}
			\vspace{3pt}
			\includegraphics[width=1.0\textwidth]{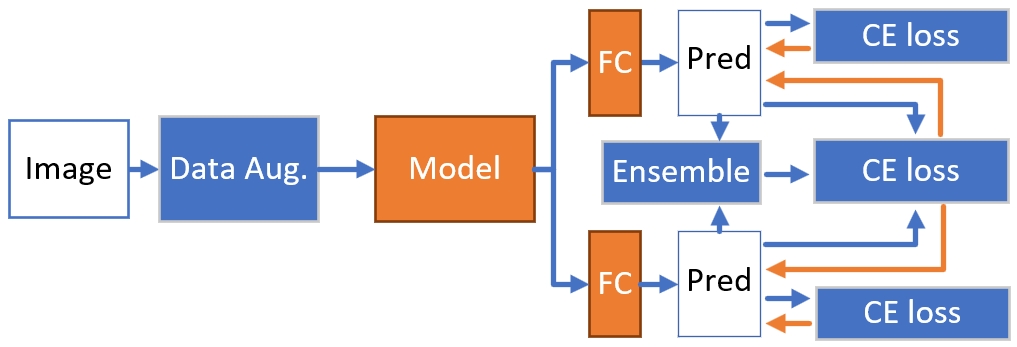}
			\centerline{(c) Multi-head Ensemble (MHE)}
		\end{minipage}
		\begin{minipage}{0.8\linewidth}
			\vspace{3pt}
			\includegraphics[width=1.0\textwidth]{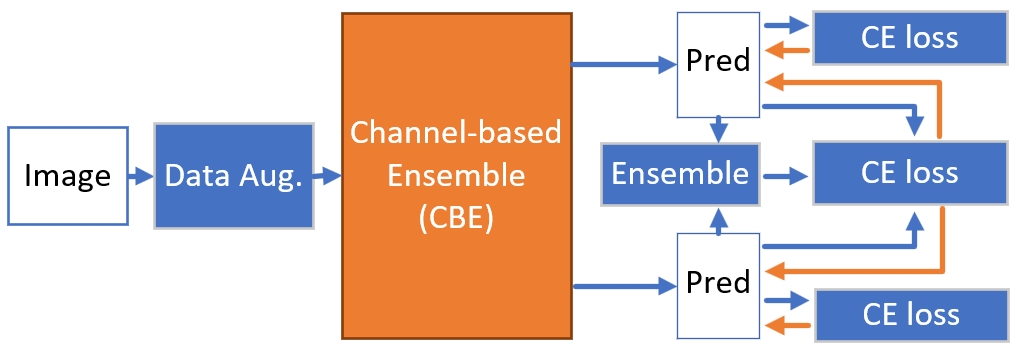}
			\centerline{(d) Our Channel-based Ensemble (CBE)}
		\end{minipage}
		\caption{The comparisons among the different ensemble methods.}
		\label{fig:Ensemble_sturcture}
	\end{figure}
	
	Fig. \ref{fig:Ensemble_sturcture}(a) shows the structure of ME methods \cite{ke2019dual, tang2021humble}. The ME uses multiple full models to construct an ensemble structure and uses the ensemble of predictions as the final output. While ME has the significant gains in terms of the ensemble performance, both the large parameter costs and the unbearable training times of ME limit its practical applications in real scenarios.
	
	Fig. \ref{fig:Ensemble_sturcture}(b) depicts the structure of the TE method \cite{laine2016temporal}. The TE utilizes exponential moving average predictions derived from the model's history to guide the training process. It can be seen as an ensemble of the historical predictions. Although TE is highly cost-effective, TE harvests a limited ensemble gain due to the lack of the theoretical guidance. 
	
	The MHE method \cite{wu2024decomposed} extends the traditional classification model predictors to a multi-head ones, as shown in Fig. \ref{fig:Ensemble_sturcture}(c). Although this method has the fewer model parameters and the shorter training time than that of ME, MHE suffers from the homogeneous prediction problems: when the number of iterations is larger than a certain number, the ensemble gain of MHE would be very limited. This issue is especially pronounced in SSL.
	
	To address the homogeneous prediction problem, we propose CBE for SSL as shown in Fig.~\ref{fig:Ensemble_sturcture}(d). Besides, We use a LB loss to maximize the un-correlations between multiple heads to avoid homogeneous prediction problem. Theoretically, CBE is a general framework to any tasks.
	
	\section{Method}
	\label{sec:Method}
	
	\subsection{Overview}
	
	\begin{figure*}[t]
		\centering
		\includegraphics[width=0.8\linewidth]{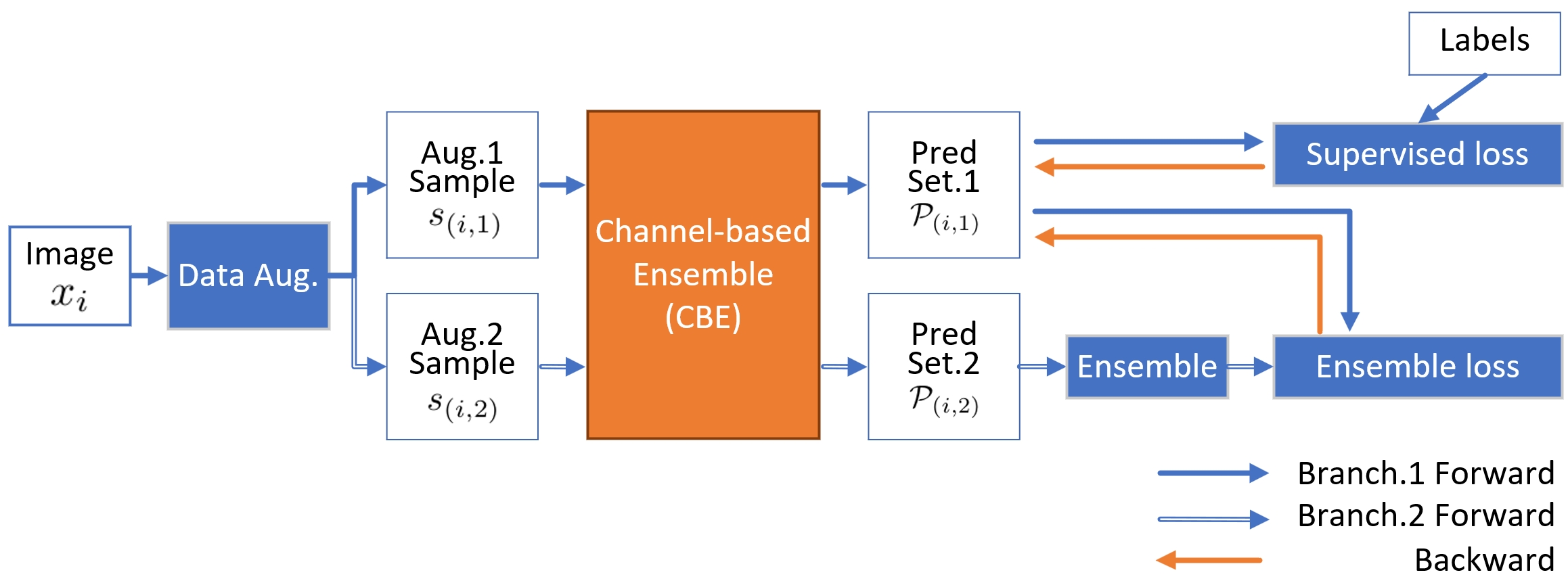}
		\caption{The neural network structure of our network.}
		\label{fig:Network_structure}
	\end{figure*}
	
	The CBE network is shown in Fig. \ref{fig:Network_structure}. First, given the data $x_i$ and the data augmentation $\omega(\cdot)$, we generate two samples $s_{(i,1)}$ and $s_{(i,2)}$ for the branches $1$ and $2$ by the data augmentation $\omega(\cdot)$. Second, the CBE generates the predicted confidence scores from each branch by the $M$ prediction heads for a sample $x_i$, i.e., $\mathcal{P}_{(i,1)}=\{p_{(i,1,m)}\}^M_{m=1}$ and $\mathcal{P}_{(i,2)}=\{p_{(i,2,m)}\}^M_{m=1}$, for $s_{(i,1)}$ and $s_{(i,2)}$, respectively. 
	
	The data augmentation strategies in the branches $1$ and $2$ can be varied according to the combined method.	For instance, when CBE is combined with FixMatch or FreeMatch, the branch $1$ employs the strong data augmentation, while the branch $2$ uses the weak data augmentation. CBE can also be combined with Mean Teacher. In this scenario, the branch $1$ and the branch $2$ can be treated as the student branch and the teacher branch, respectively. 

	
	For the labeled data $\{(x_i,y_i)\}^{N_L}_{i=1}$, the supervised training is supervised by the ground truth $y_i$ as follows:
	\begin{small}
		\begin{equation}\label{eqt:SupevisedLoss}	L_{l}=\frac{1}{N_B}\sum_{i=1}^{N_B}\frac{1}{M}\sum_{m=1}^{M}\frac{1}{2}[\text{CE}\big(p_{(i,1,m)}, y_{i}\big)+\text{CE}\big(p_{(i,2,m)}, y_{i}\big)] ,
		\end{equation}
	\end{small}
	where $N_B$ is the batch size of the labeled data, and $\text{CE}(\cdot,\cdot)$ is the cross-entropy function. 
	
	For unlabeled data $\{x_i\}^{N_L+N_U}_{i=N_L+1}$, the prediction scores $\mathcal{P}_{(i,2)}$ is used to filter the unreliable prediction. The ensemble prediction $\overline{\mathcal{P}}_{(i,2)}$ is as follows:
	\begin{equation}\label{eqt:PseudoLabel}
		\overline{\mathcal{P}}_{(i,2)}=\frac{1}{M}\sum_{m=1}^{M}\mathcal{T}(\text{max}(p_{(i, 2, m)}) >\tau)\cdot p_{(i, 2, m)},
	\end{equation}
	where $\mathcal{T}(\cdot>\tau)$ is the indicator function for the confidence-based thresholding in which $\tau$ is the threshold.
	
	$\overline{\mathcal{P}}_{(i,2)}$ in~\eqref{eqt:PseudoLabel} is further used as a PL to supervise the $\mathcal{P}_{(i,1)}$. Consequently, the ensemble supervised loss $L_{e}$ is as follows:
	
	\begin{equation}\label{eqt:EnsembleLoss}
		L_{e}=\frac{1}{\mu N_B}\sum_{i=1}^{\mu N_B}\frac{1}{M}\sum_{m=1}^{M}\text{CE}\big(p_{(i,1,m)},\overline{\mathcal{P}}_{(i,2)}\big),
	\end{equation}
	where $\mu$ is the ratio of the unlabeled data number to the labeled data one.
	
	As discussed in Sec.~\ref{sec:related-ensemble}, the CBE faces the homogeneity problem during the training process, which eventually losses the advantages of the ensemble learning. This problem is fatefully harmful to SSL. Because the empirical experiments (as will be discussed in Sec.~\ref{sec:exp-qualityof-label}) show that only a few or even no samples obtain the correct PLs in each iteration. If the homogeneity problem occurs, CBE would turn into a single model-based method.     
	
	\subsection{Chebyshev Constraint}
	
	Ensemble learning effectively mitigates the prediction bias;  however, the predictions from each predictor tend to be homogeneous. We theoretically fund the Chebyshev constraint to solution this problem.
	
	Let $\mathcal{P}_{i}=\{p_{(i,m)}\}^M_{m=1}$ be the predictions for the unlabeled sample $s_{i}$ from data $x_i$ by the $M$ prediction heads. 	
	
	\textbf{Lemma 1.} The theoretical error $\mathcal{E}_{i}$ between the ensemble prediction $\overline{\mathcal{P}}_{i}$ and the theoretical ground truth $\mathcal{P}_{i}^{*}$ for the sample $x_i$ is computed as follows:
	\begin{small}
		\begin{equation}\label{eqt:Ensemble}
			\begin{aligned}
				\mathcal{E}_{i} &= P\{|\overline{\mathcal{P}}_{i} - \mathcal{P}_{i}^{*}|\} \\
				&= P\{|\mathcal{P}_{i}^{*} - \mathbb{E}(\mathcal{P}_{i}^{*})| \ge \epsilon \} \\
				&\le \frac{1}{\epsilon^2}\text{var}(\mathcal{P}_{i}^{*}) \\
				&\le \frac{1}{\epsilon^2}\frac{1}{M^2} \big[\underbrace{\sum_{m=1}^{M}\text{var}(p_{(i,m)})}_{\text{var}(\mathcal{P}_{i})} + \underbrace{\sum_{m=1}^{M}\sum_{j=1,j\neq m}^{M}2\text{covar}(p_{(i,m)},p_{(i,j)})}_{\text{covar}(\mathcal{P}_{i})}\big], \\
			\end{aligned}
		\end{equation}
	\end{small}
	where $\epsilon$ is any positive number,  $\mathbb{E}(\cdot)$ is the expectation, and $ \text{var}(\cdot) $, $ \text{covar}(\cdot) $ are the functions of variance and covariance, respectively.
	
	\textbf{Lemma 2.} Assuming that the multi-head predictor predicts the data $x_i$ $K$ times, the variance $\mathcal{V}_{i}$ of the $K$ ensemble predictions can be expressed as:
	\begin{equation}\label{eqt:Variance}
		\begin{aligned}
			\mathcal{V}_{i} &= \frac{1}{K}\sum_{k=1}^{K}\big(\frac{1}{M}\sum_{m=1}^{M}p_{(i,m,k)}-\frac{1}{M}\sum_{m=1}^{M}\mathbb{E}(\mathcal{P}_{(i,m)})\big) \\
			&\le\underbrace{\frac{1}{M}\sum_{m=1}^{M}\text{var}(\mathcal{P}_{(i,m)})}_{\overline{\text{var}}(\mathcal{P}_{(i,m)})}+\frac{1}{M}\underbrace{\frac{1}{K}\sum_{k=1}^{K}\text{covar}(\mathcal{P}_{(i,k)})}_{\overline{\text{covar}}(\mathcal{P}_{(i,k)})}, \\
		\end{aligned}
	\end{equation}
	
	where $ \text{var}(\cdot) $, $ \text{covar}(\cdot) $ are the functions of variance and covariance, respectively. $p_{(i,m,k)}$ is the $k$ prediction of $m$-th head predictor for the data $x_i$, and $\mathbb{E}(\mathcal{P}_{(i,m)})$ is the expectation of $K$ prediction of $m$-th head predictor, $\mathcal{P}_{(i,k)}$ is the predictions of all head predictors at $k$ time, $\overline{\text{var}}(\mathcal{P}_{(i,m)})$ is the average variance of $K$ prediction of all head, and $\overline{\text{covar}}(\mathcal{P}_{(i,k)})$ is the average covariance of the $K$ time predictions for each head. 
	
	Since $\overline{\text{covar}}(\mathcal{P}_{(i,k)})$ is a very small value, ~\eqref{eqt:Variance} indicates that the variance of the ensemble prediction is not greater than the prediction variance of each prediction head.
	
	\textbf{Remark.} The factors affecting the variance of the ensemble prediction error in~\eqref{eqt:Variance} include the stability of each predictor ($\text{var}(i.e., \mathcal{P}_{i})$) and the diversity of ensemble (i.e., $\text{covar}(\mathcal{P}_{i})$).  Note that the variance of a prediction is especially important to SSL. Because the changeable PLs would hamper SSL to efficiently harvest correct PLs during the training process.
	
	(a) \textbf{Stability} The variance $\text{var}(i.e., \mathcal{P}_{i})$ in~\eqref{eqt:Ensemble} represents the sensitivity of a predictor to the different data augmentation (or perturbation) for the same sample. As a result, smaller variance value in~\eqref{eqt:Variance} implies that a predictor is stable, which is consistent with~\eqref{eqt:Ensemble}.
	
	(b) \textbf{Diversity} Correlations among the multiple predictors should tend to be zeros.~\eqref{eqt:Ensemble} indicates that the lower the correlation between the predictors are, the smaller the ensemble prediction error is; This is consistent with~\eqref{eqt:Ensemble}.
	
	(c) \textbf{Variance} ~\eqref{eqt:Variance} shows that the ensemble method in~\eqref{eqt:Ensemble} can effectively reduce the variance of the predictions.
	
	\subsection{Low Bias Loss}
	
	\begin{figure}[t!]
		\centering
		\includegraphics[width=0.8\linewidth]{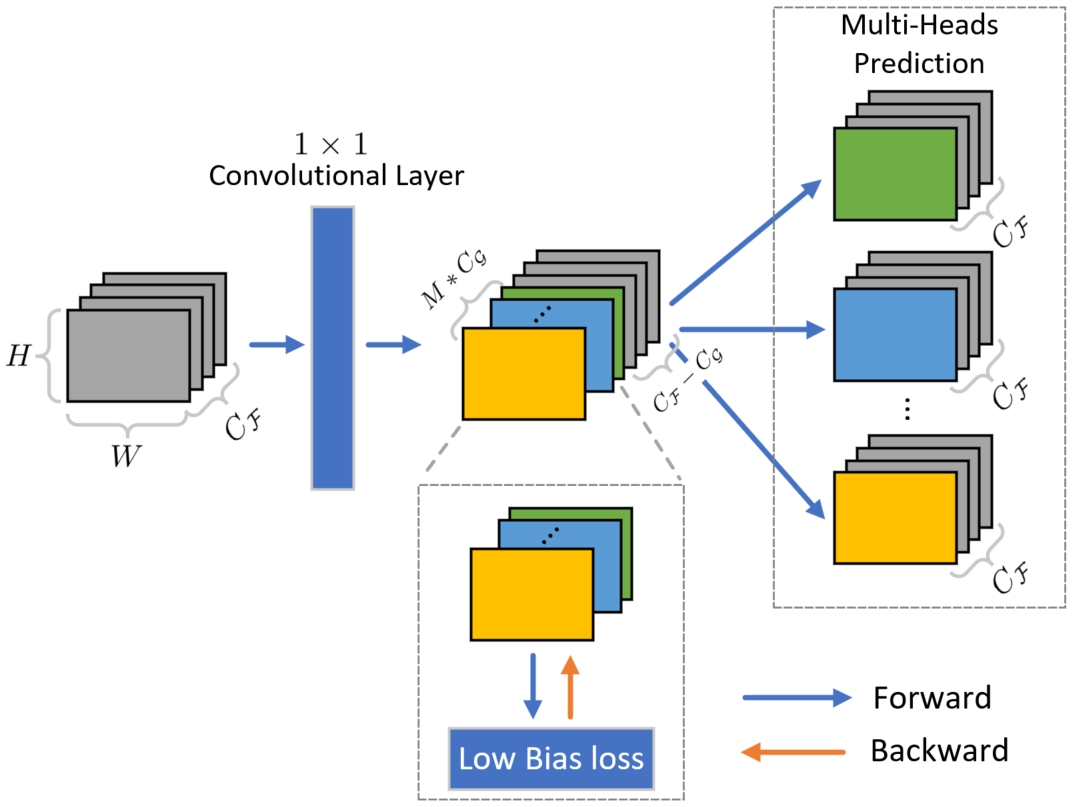}
		\caption{The structure of Low Bias loss (LB loss).}
		\label{fig:Low_Bias_Loss}
	\end{figure}
	
	In CBE, as shown in Fig. \ref{fig:Low_Bias_Loss}, the $1\times1$ convolutional kernel expands the original feature $\mathcal{F}$ from $C_{\mathcal{F}}\times H\times W$ to $[C_{\mathcal{F}}+(M-1)*C_{\mathcal{G}}]\times H\times W$, where $M$ is the head number of the multi-head predictors, $C_{\mathcal{F}}$ and $C_{\mathcal{G}}$ are the channels number of the original feature and the private feature $\mathcal{G}$ for each head, respectively. That is, the expanded feature is split into the $M$ sub-features. Each sub-feature for a prediction heard is consist of $C_\mathcal{F}$ and $C_\mathcal{G}$, where $C_\mathcal{F}$ is shared among the $M$ prediction heads.The CBE empirically has the negligible increase of computation cost, compared with a single model
	
	Based on the~\eqref{eqt:Ensemble} and the structure of CBE in~\ref{fig:Low_Bias_Loss}, we propose a Low Bias (LB) loss function to minimize the correlation among each head as follows:
	
	\begin{equation}\label{eqt:FULoss}
		L_{fu}=\frac{1}{\mu N_B}\sum_{i=1}^{\mu N_B}\frac{1}{M}\sum_{i=1}^{M}\sum_{j=1,j\neq i}^{M}\text{COV}(\mathcal{G}_{i},\mathcal{G}_{j}),
	\end{equation}
	where $ \text{COV}(\mathcal{G}_{i},\mathcal{G}_{j}) $ is the correlation function between the features of two heads. $\mathcal{G}_{i}$ is the private feature fro each prediction head.
	
	\subsection{Low Variance Loss}
	
	\begin{table*}[t]
		\centering
		\begin{tabular}{@{}lcccccc@{}}
			\toprule 
			\multirow{2}{*}{Method} & \multicolumn{3}{c}{CIFAR10} & \multicolumn{3}{c}{CIFAR100}\\
			& 40 & 250 & 4000 & 400 & 2500 & 10000 \\
			\midrule
			FixMatch \cite{sohn2020fixmatch} & 8.15 & 5.96 & 5.05 & 51.74 & 30.09 & 22.69 \\
			FixMatch + CBE & \pmb{7.21} & \pmb{6.88} & \pmb{4.63} & \pmb{51.17} & \pmb{29.50} & \pmb{22.60} \\
			FreeMatch \cite{wang2022freematch} & 14.85 & 5.85 & 4.95 & 44.41 & 28.04 & 22.37 \\
			FreeMatch + CBE & \pmb{6.13} & \pmb{5.25} & \pmb{4.55} & \pmb{43.64} & \pmb{26.85} & \pmb{22.33} \\
			\bottomrule
		\end{tabular}
		\caption{The error rates of the SSL classification on CIFAR10, CIFAR100 datasets. 'FixMatch' and 'FreeMatch' denote the use of them only, respectively. 'X+ CBE' denotes the combination of X and our CBE.}
		\label{tab:Comparisons}
	\end{table*}
	
	We propose a Low Variance (LV) loss function to reduce the variance of the predicted distribution for the unlabeled data by utilizing the ground truth of the labeled data as constraints. 
	
	The variance $\mathcal{V}_{B}$ of the predictions of the multi-head predictor for a batch data can be represented by the variance between the ensemble prediction and the ground truth as follows:
	\begin{equation}\label{eqt:Variance_LV}
		\begin{aligned}
			\mathcal{V}_{B} &= \text{var}(\overline{\mathcal{P}}_{B} - \mathcal{P}_{B}^{*}) \\
			&= \text{var}(\overline{\mathcal{P}}_{B}) + \text{var}(\mathcal{P}_{B}^{*}) - 2\text{covar}(\overline{\mathcal{P}}_{B},\mathcal{P}_{B}^{*}), \\
		\end{aligned}
	\end{equation}
	where $\overline{\mathcal{P}}_{B}$ and $\mathcal{P}^{*}_{B}$ are the arrays of the ensemble predictions and the ground truth values of the labeled data in a batch, respectively.
	
	~\eqref{eqt:Variance_LV} indicates that maximizing the $\text{covar}(\overline{\mathcal{P}}_{B},\mathcal{P}_{B}^{*})$ between the ensemble prediction $\overline{\mathcal{P}}_{B}$ and ground truth $\mathcal{P}_{B}^{*}$ could reduce the variance values. Therefore, the LV loss $L_{lv}$ is approximately computed as follows: 
	
	\begin{equation}\label{eqt:LVLoss}
		L_{lv}=1-\text{COV}(\overline{\mathcal{P}}_{B},\mathcal{P}^{*}_{B}))
	\end{equation}
	
	
	\subsection{Total Loss}
	
	The overall objective for CBE is:
	
	\begin{equation}\label{eqt:TotalLoss}
		L=\lambda_{l}L_{l}+\lambda_{e}L_{e}+\lambda_{fu}L_{fu}+\lambda_{lv}L_{lv} ,
	\end{equation}
	where $\lambda_{l}$, $\lambda_{e}$, $\lambda_{fu}$, $\lambda_{lv}$ are the balanced parameters for $L_{l}$, $L_{e}$, $L_{fu}$, $L_{lv}$, respectively. In this paper, all the weights $\lambda_{u}$, $\lambda_{ens}$, $\lambda_{fd}$, $\lambda_{mc}$ are simply set to 1.
	
	\subsection{Implementation Details}

	For the PL method, PLs are used as the supervised signal, and the cross-entropy loss function is used to drive unsupervised learning. To accommodate various SSL methods, our approach incorporates the threshold setting strategy corresponding to the original SSL methods. That is, We replace the unsupervised loss function in original SSL methods with the proposed ensemble loss function in~\eqref{eqt:EnsembleLoss}.
	
	\section{Experiments}
	\label{sec:Experiments} 
	
	We evaluate the efficacy of our approach on the CIFAR10/100 dataset \cite{krizhevsky2009learning} to show the improvement of the classical SSL methods combined with our CBE. FixMatch and FreeMatch are chosen as the SOTA SSL approaches. Concretely, FixMatch is a classical weak-strong approach, while FreeMatch represents the SOTA in the current SSL techniques. It is important to note that when these SSL methods combined with our approach, it only requires three modifications: 1) Using the CBE class library provided by us to modify the classification model into a multi-head prediction model;  2) Employing the threshold strategy from the original SSL method to the ensemble predictions for generating PLs; and 3) Replacing the unsupervised loss function from the original SSL method with the ensemble supervised loss function~\eqref{eqt:EnsembleLoss}. 
	
	\subsection{Configurations}
	The classification model used in all experiments is Wide ResNet (WRN) with a widen-factor of 2 and 6 for CIFAR-10 and CIFAR-100 datasets, respectively.
	
	For a fair comparison, the same hyper-parameters are used for all experiments. Specifically, all experiments are performed with an standard Stochastic Gradient Descent (SGD) optimizer with momentum 0.9 \cite{sutskever2013importance, polyak1964some} and Nesterov momentum \cite{dozat2016incorporating} enabled.The learning rate is 0.03. The batch size of the labeled data is 32, the batch size of the unlabeled data is 224 (i.e., $ \mu=7 $). We used the same decay value (i.e., 0.999) for the experiment involving the Exponential Moving Average (EMA) \cite{tarvainen2017mean}. Data augmentation on labeled data includes random horizontal flipping and random cropping. For the unlabeled data, weak data augmentation is similar, but strong data augmentation adds RandAugment to it \cite{cubuk2020randaugment}.
	
	The confidence thresholds $ \tau $ are 0.95, 0.9 for FixMatch and our method, respectively. FreeMatch threshold is dynamically controlled by the self-adaptive thresholding (SAT) technique. The number of multi-heads in CBE is 5. 
	
	Due to the hardware limitations, all experiments were trained for 200 epochs with 1024 iterations within each epoch. We used a fixed random seed (1388) to obtain accurate and reliable experimental results.
	
	\subsection{Comparisons}
	
	The Top1 error rates for the CIFAR-10 and CIFAR-100 datasets are shown in Table \ref{tab:Comparisons}. It is worth noting that CBE consistently outperforms the remaining methods by significant margins in the case of the limited labeled data. For example, for CIFAR-10, CBE with 40 labels improves FixMatch and FreeMatch by 0.94\% and 8.72\%, respectively; For CIFAR-100, FixMatch and FreeMatch are improved by 0.67\% and 0.77\%, respectively when  400 labeled data are used. These improvements demonstrate the power and potential of the proposed CBE in SSL for reducing the bias and variance of the pseudo-labels in the ST process.
	
	The existing SSL methods primarily focus on formulating threshold policies, while overlook the intrinsic characteristics of the PLs themselves. For instance, the superior performance of FreeMatch in comparison to FixMatch can be seemingly attributed to the enhancement of its threshold policy. On the contrast, our CBE method effectively enhances both FixMatch and FreeMatch by generating the unbiased and low-variance PLs. The results further indicate the dynamical characteristics of the PLs is critical to SSL. 
	
	\subsection{Ablation}
	
	\begin{table}[h]
		\centering
		\begin{tabular}{@{}lc@{}}
			\toprule 
			Method & CIFAR10@40 \\
			\midrule
			FreeMatch \cite{wang2022freematch}                    & 14.85       \\
			\midrule
			FreeMatch + CBE (no LB and LV loss)       & 10.26       \\   
			FreeMatch + CBE (no LV loss)                          & 8.83       \\ 
			FreeMatch + CBE                                       & \pmb{6.13}  \\
			\bottomrule
		\end{tabular}
		\caption{Results of the ablation experiments on CIFAR-10.}
		\label{tab:Ablation}
	\end{table} 
	
	\begin{figure}[h!]
		\centering
		\includegraphics[width=0.8\linewidth]{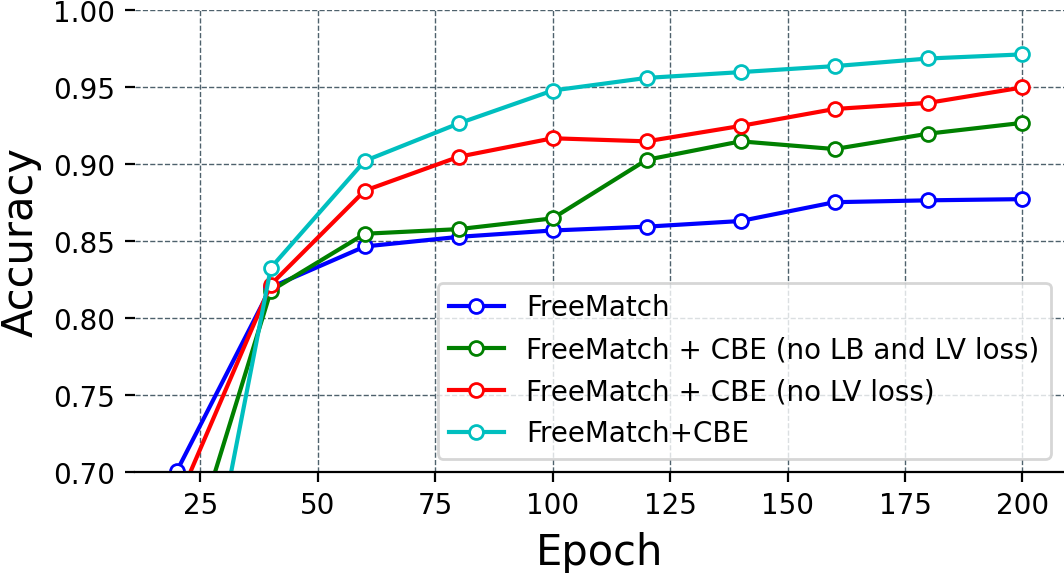}
		\caption{Accuracy curves of PLs in ablation.}
		\label{fig:Ablation_Accuracy}
	\end{figure}
	
	\begin{figure*}[t!]
		\begin{minipage}{0.33\linewidth}
			\vspace{3pt}
			\includegraphics[width=\textwidth]{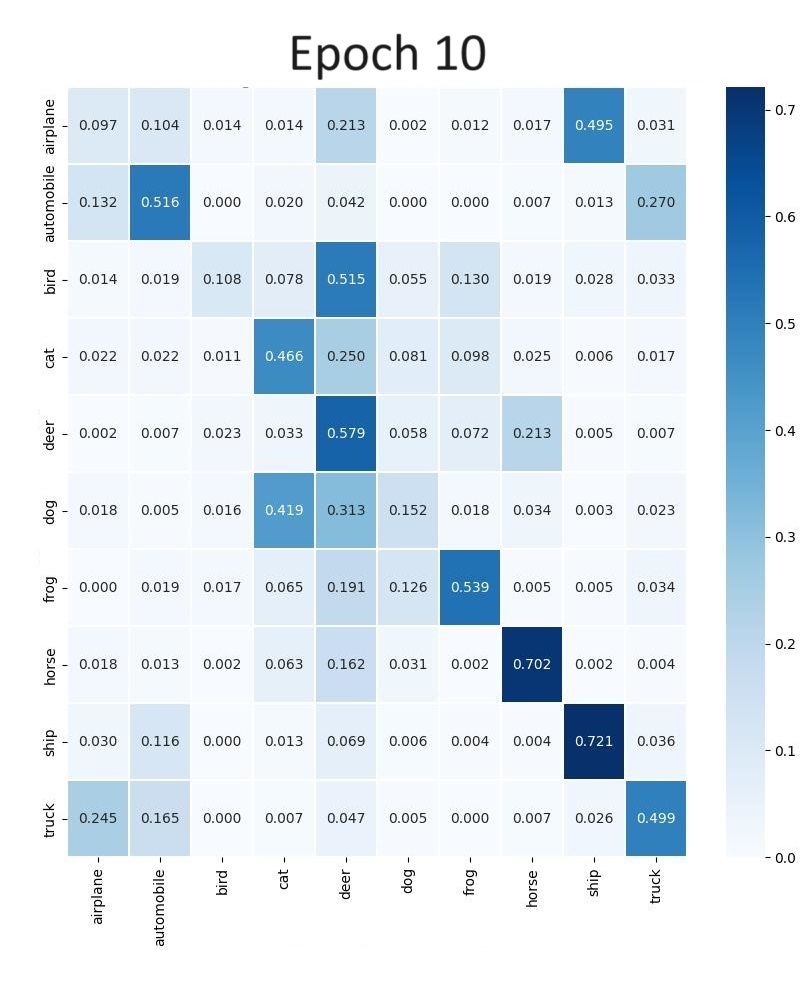}
		\end{minipage}
		\begin{minipage}{0.33\linewidth}
			\vspace{3pt}
			\includegraphics[width=\textwidth]{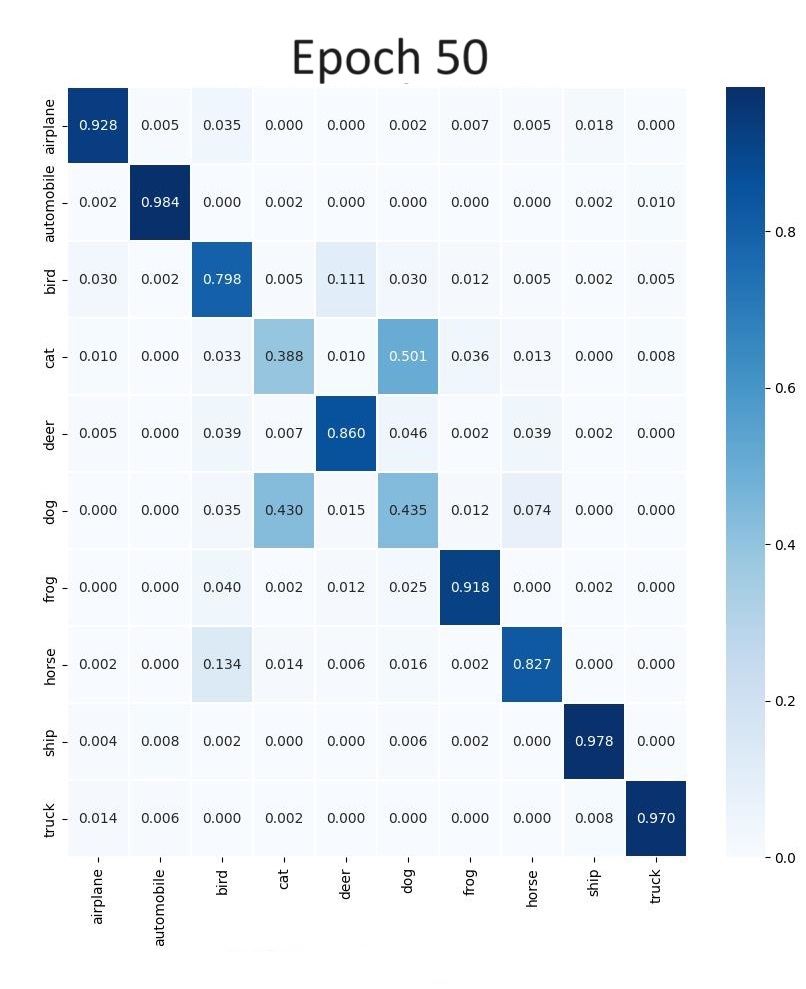}
		\end{minipage}
		\begin{minipage}{0.33\linewidth}
			\vspace{3pt}
			\includegraphics[width=\textwidth]{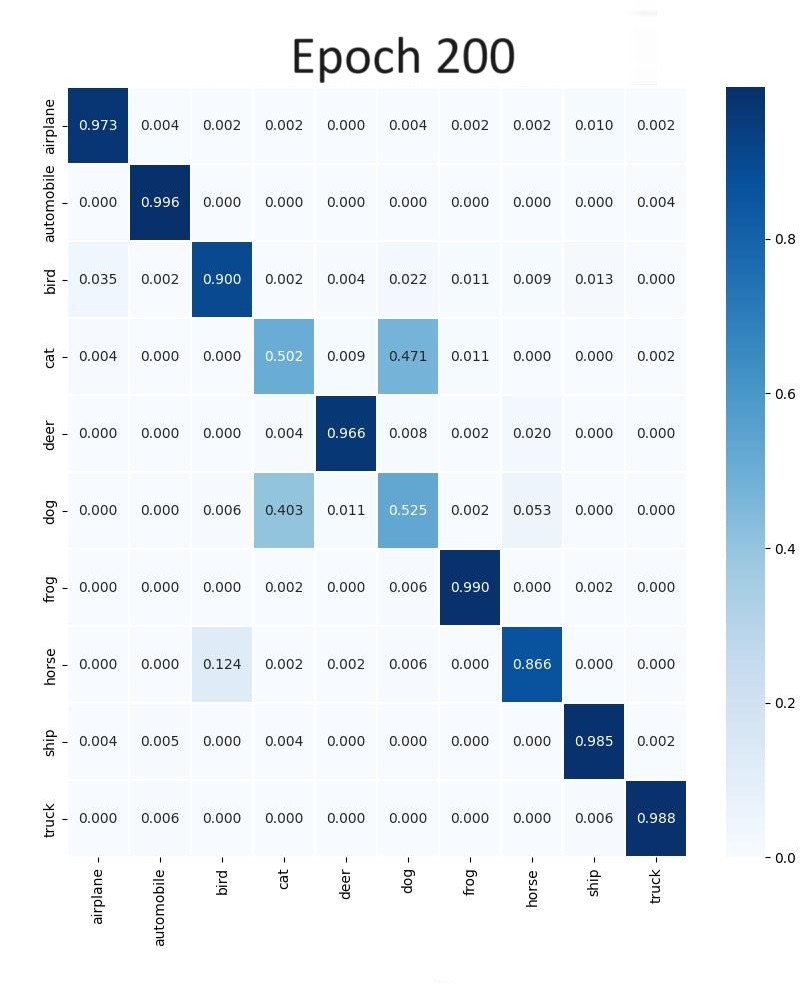}
		\end{minipage}
		\centerline{(a) Confusion matrix of FreeMatch}
		\begin{minipage}{0.33\linewidth}
			\vspace{3pt}
			\includegraphics[width=\textwidth]{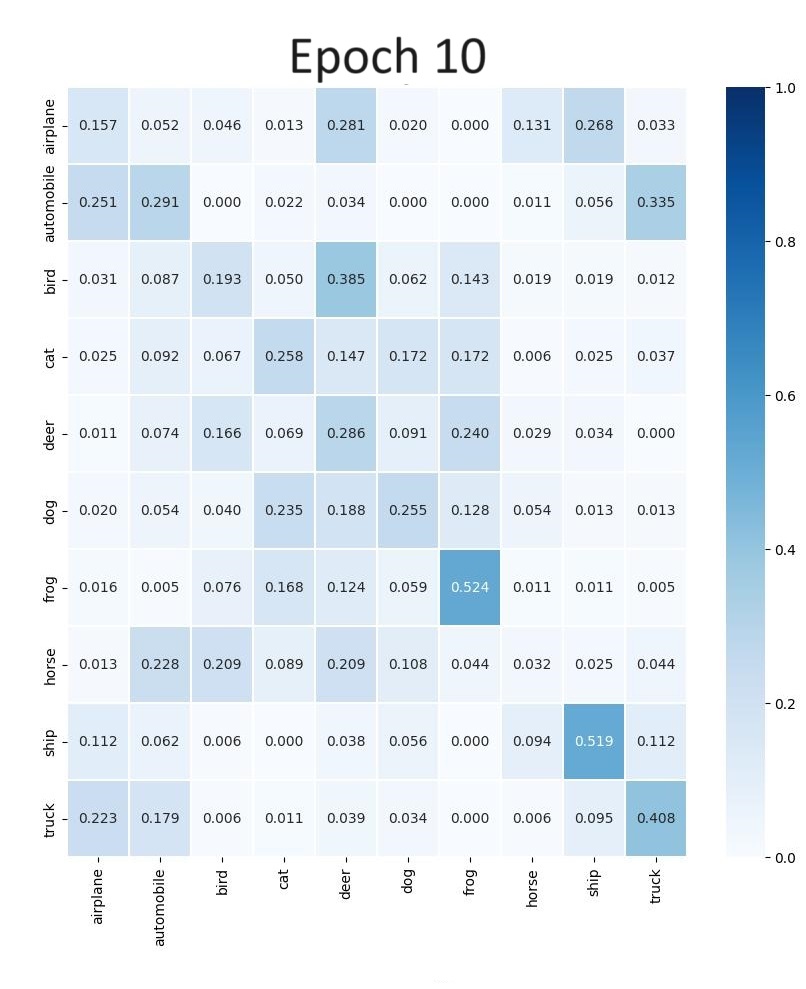}
		\end{minipage}
		\begin{minipage}{0.33\linewidth}
			\vspace{3pt}
			\includegraphics[width=\textwidth]{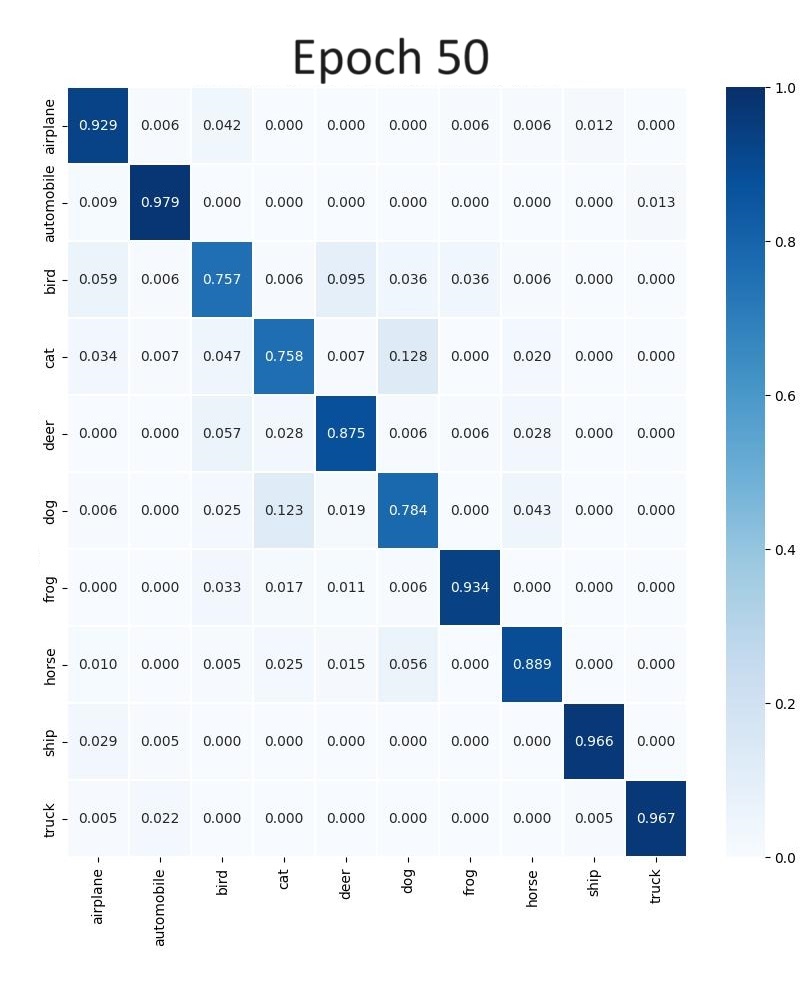}
		\end{minipage}
		\begin{minipage}{0.33\linewidth}
			\vspace{3pt}
			\includegraphics[width=\textwidth]{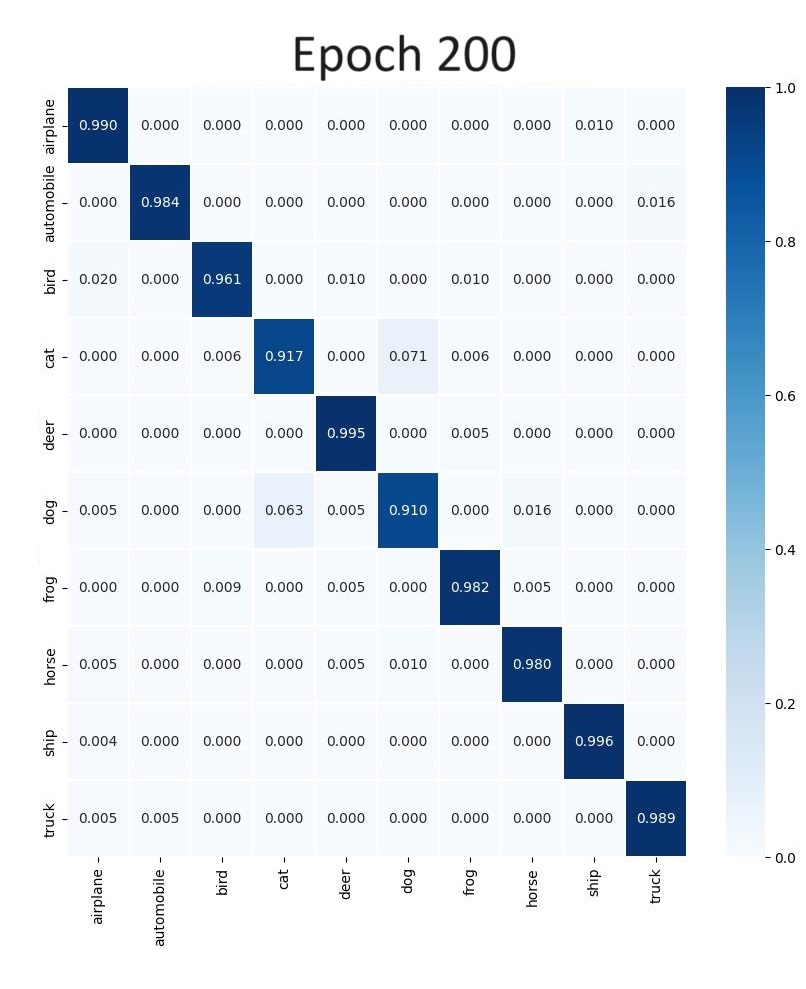}
		\end{minipage}
		\centerline{(b) Confusion matrix of FreeMatch+CBE}
		\caption{Comparison the confusion matrix of FreeMatch in different epoch.}
		\label{fig:PLs_confusion_matrix}
	\end{figure*}
	
	The results of the ablation experiments are shown in Table.~\ref{tab:Ablation}. An experiment  with 40 labeled data on CIFAR-10 dataset is used as an example to evaluate the role of each module in CBE.
	
	Firstly, only the network structure of CBE in Fig.~\ref{fig:Network_structure} is used, while both the LB loss in~\eqref{eqt:FULoss} and the LV loss~\eqref{eqt:LVLoss} are removed. We denote this setting as FreeMatch + CBE (no LB and LV loss), which improves FreeMatch by 4.59\% by constructing an ensemble prediction mechanism based on the multiple prediction heads. The improvement indicates that the ensemble structure indeed helps SSL methods to reduce the bias and variance of the PLs. However, as shown in Fig.~\ref{fig:Ablation_Accuracy}, using only CBE without LB and LV losses tends to make the predictions of multi-head predictors homogeneous, which compromises the accuracy of pseudo-labels.
	
	Secondly, when both the network structure of CBE and the LB loss are adopted, we denote this setting as FreeMatch + CBE (no LV loss). FreeMatch + CBE (no LV loss) performance is improved by 1.43\%, compared with the FreeMatch + CBE (no LB and LV loss). This indicates that the un-correlations among the predictors in~\eqref{eqt:Ensemble} is an important factor to bring the ensemble gain. As a result, FreeMatch + CBE (no LB and LV loss) achieves a better performance than FreeMatch, benefiting for the further improvement in the quality of PLs.
	
	Finally, as shown in Table.~\ref{tab:Ablation}, when enabling the LV loss, based on the FreeMatch + CBE (no LV loss), we denote this setting as FreeMatch + CBE. FreeMatch + CBE performance is improved by 2.20\%, compared with the FreeMatch + CBE (no LV loss). This indicates that the un-correlations between the ensemble prediction and the ground truth is an important factor to reduce the variance of predictions, which is shown in~\eqref{eqt:Variance_LV}.
	
	\subsection{Quality of Pseudo-Labels}\label{sec:exp-qualityof-label}
	
	The Sampling Rate (SR)~\cite{sohn2020fixmatch,wang2022freematch} of PLs is an important factor to affect the accuracy of the final model due to the ST process in SSL. If more number of PLs are used, richer knowledge from sampled data and the corresponding PLs is supplied to a model~\cite{wang2022freematch}. However, this assumption faces the problem caused by the wrongly predicted PLs. Therefore, We use SR to calculate the quality of PLs when two models have the same accuracy at the same iteration. Specially, the SR~\cite{wang2022freematch} is computed as follows:
\begin{equation}\label{eqt:Sampling_rate}
	\eta=\frac{1}{\mu N_B}\sum_{i=1}^{\mu N_B}\mathcal{T}(\text{max}(p_{i}) >\tau),
\end{equation}
where $p_{i}$ is the prediction of a sample $x_i$, and $\mathcal{T}(\cdot>\tau)$ is the indicator function in which $\tau$ is the threshold. 
For CBE, the sampling of PLs is based on the ensemble prediction. Therefore, we define the CBE SR $\eta_{CBE}$ of PLs as follow:
\begin{equation}\label{eqt:CBE_sampling_rate}
	\eta_{CBE}=\frac{1}{\mu N_B}\sum_{i=1}^{\mu N_B}\Gamma(\frac{\sum_{m=1}^{M}\mathcal{T}(\text{max}(p_{(i, m)}) >\tau)}{M} >\gamma),
\end{equation}
where $p_{(i, m)}$ is the prediction of the $m$-th head of the multi-head predictors for the sample $x_i$, and $\Gamma(\cdot>\gamma)$ is the indicator function in which $\gamma$ is the ensemble sampling threshold.

The accuracy of PLs and SR curves of each SSL method are shown in Fig.~\ref{fig:PLs_Quality}(a) and Fig.~\ref{fig:PLs_Quality}(b), respectively. If we Compare FixMatch and FreeMatch, it can be seen that FreeMatch outperforms FixMatch by setting a lower and more reasonable confidence threshold and increasing the pseudo-label SR. Interestingly, CBE significantly surpasses FreeMatch by improves the accuracy of PLs while reducing SR of PLs. This shows that CBE guarantees the richness of knowledge in PLs by improving the correctness of the knowledge used for the ST process.

\begin{table}[t!]
	\centering
	\begin{tabular}{@{}lcc@{}}
		\toprule 
		Method & Training time (minutes) & CIFAR10@40 \\
		\midrule
		FixMatch      & 4096 & 7.47 \\   
		FixMatch + CBE & 2048  & \pmb{7.45} \\ 
		FixMatch + CBE & 4096 & \pmb{5.20} \\
		\bottomrule
	\end{tabular}
	\caption{Error rates of FixMatch and FixMatch+CBE on the CIFAR10@40 dataset for different training time. The training time is measured on a GPU3090.}
	\label{tab:Effectiveness_and_efficiency}
\end{table}

FixMatch and FreeMatch mainly improve the mechanism of the PL filtering and ignore the impacts of the noise from PLs, which reduces the training efficiency. In contrast, as shown in Table.~\ref{tab:Effectiveness_and_efficiency}, CBE significantly outperforms FixMatch by only using about $1/2$ training time of the FixMtach. This comparative results indicates the importance of the low-bias and low variance PLs is critical to SSL.

Moreover, the quality of the PLs is further illustrated as the confusion matrix in Fig. \ref{fig:PLs_confusion_matrix}, our CBE method outperforms FreeMatch in terms of the PL confusion matrices when trained on the CIFAR10@40 dataset. Conversely, compared with CBE, FreeMatch fails to generate more accurate PL. Because FreeMatch has a low accuracy for the early training stage PLs (i.e., both \#10 and \#50 iterations); besides, the wrong PLs for certain samples tend to confuse the model in the later training stage.

\subsection{The Computational Cost of CBE}

\begin{table}[h]
	\centering
	\begin{tabular}{@{}lcc@{}}
		\toprule 
		Method & Model Parameters & FLOPs \\
		\midrule
		FixMatch/FreeMatch      & 216.254M & 1.468M \\   
		FixMatch/FreeMatch + CBE & 216.390M & 1.473M \\ 
		\bottomrule
	\end{tabular}
	\caption{The comparison of the computational cost of FixMatch, FreeMatch, and CBE.}
	\label{tab:Cost}
\end{table}

The advantage of this approach is that it can efficiently implement ensemble predictions. As shown in Table. \ref{tab:Cost}, when both FixMatch and FreeMatch are combined with the proposed CBE, CBE only adds 0.136M model parameters and increases 0.005M FLOPs. Compared to the traditional ensemble methods, our approach provides the superior performance by reducing both the bias and variance of the PLs with a negligible computational cost.

\section{Conclusions}
\label{sec:Conclusions}

To our best knowledge, this paper firstly discussed that PL methods, influenced by the characteristics of self-trained models, tend to generate biased and high-variance predictions due to the accumulation of their own errors, especially when the labeled data is limited. Although ensemble learning can alleviate this issue, existing ensemble methods have inherent limitations and are not well-suited for SSL. To address this concern, we propose a lightweight and efficient ensemble approach called Channel-based Ensemble (CBE) to provide unbiased and low-variance pseudo-labels for SSL. Our proposed method incorporates the Chebyshev constraint: a Low Bias loss that maximizes the feature differences across multiple heads to maintain ensemble diversity and reduce prediction bias. as well as a Low Variance loss that encourages higher standardized probability distributions in order to decrease prediction variance. Extensive experiments were conducted to demonstrate the effectiveness of our proposed method. Moreover, our approach can be easily extended to any SSL method such as FixMatch or FreeMatch.

\nocite{langley00}

\bibliography{example_paper}
\bibliographystyle{icml2024}

\newpage
\appendix
\onecolumn

\end{document}